\documentclass[a4paper]{article}
\usepackage{cite}

\usepackage{subfigure}
\usepackage{mathrsfs}
\usepackage{float}
\usepackage{hyperref}
\usepackage{multirow}
\usepackage{graphicx}
\usepackage{xcolor}
\usepackage{amsmath}
\usepackage{booktabs}
\usepackage{threeparttable}
\usepackage[nodisplayskipstretch]{setspace}

\usepackage{hyperref}
\usepackage{xurl}

\usepackage{INTERSPEECH2021}
\title{Limited Data Emotional Voice Conversion Leveraging Text-to-Speech: \\ Two-stage Sequence-to-Sequence Training}
\name{Kun Zhou$^*$, \thanks{\textbf{Codes \& Speech Samples}: \url{https://kunzhou9646.github.io/IS21/}}Berrak Sisman$^{\dag}$, Haizhou Li$^*$}
\address{
  $^*$Department of ECE, National University of Singapore (NUS), Singapore\\
  $^\dag$Singapore University of Technology and Design (SUTD), Singapore}
\email{zhoukun@u.nus.edu, berraksisman@u.nus.edu, haizhou.li@nus.edu.sg}

\begin{document}
\sloppy
\setlength{\lineskiplimit}{0pt}
\setlength{\lineskip}{0pt}
\setlength{\abovedisplayskip}{0pt}
\setlength{\belowdisplayskip}{0pt}
\setlength{\abovedisplayshortskip}{0pt}
\setlength{\belowdisplayshortskip}{0pt}    

\maketitle
\begin{abstract}
Emotional voice conversion (EVC) aims to change the emotional state of an utterance while preserving the linguistic content and speaker identity. In this paper, we propose a novel 2-stage training strategy for sequence-to-sequence emotional voice conversion with a limited amount of emotional speech data. We note that the proposed EVC framework leverages text-to-speech (TTS) as they share a common goal that is to generate high-quality expressive voice. In stage 1, we perform style initialization with a multi-speaker TTS corpus, to disentangle speaking style and linguistic content. In stage 2, we perform emotion training with a limited amount of emotional speech data, to learn how to disentangle emotional style and linguistic information from the speech. The proposed framework can perform both spectrum and prosody conversion and achieves significant improvement over the state-of-the-art baselines in both objective and subjective evaluation.

%We also propose a joint training between emotional voice conversion and emotional text-to-speech systems without the need for parallel data.
%We first pre-train the proposed framework with a large text-to-speech corpus, where the style encoder learns style representation from the speech. 
%We then train the framework with a limited amount of emotional speech data, where the style encoder acts as an emotion encoder to learn emotional style representation. The proposed framework is capable to perform both spectrum and prosody conversion, and achieves significant improvement over the state-of-the-art  baselines.
\end{abstract}
\vspace{-3mm}
\section{Introduction}
Sequence-to-sequence (seq2seq) speech synthesis frameworks, such as Tacotron \cite{wang2017tacotron}, can generate high-quality synthetic speech. However, such frameworks heavily rely on a large amount of training data. Furthermore, they generally lack emotional variance~\cite{schuller2018age}. Emotional voice conversion aims to convert the emotional state of speech from one to another while preserving the linguistic content and speaker identity. This technique allows us to project a desired emotion into the generated speech, thus bears huge potential in real-world applications, such as expressive text-to-speech~\cite{liu2020expressive}.

%Both of them aim to convert para-linguistic information while maintaining linguistic content~\cite{sisman2020overview}. 

Emotion is inherently supra-segmental and complex with multiple signal attributes concerning both spectrum and prosody~\cite{xu2011speech}, thus it is insufficient to convert the emotion only with frame-wise spectral mapping. Prosodic features, such as pitch, energy, and duration, also need to be dealt with for emotional voice conversion. We believe that seq2seq training is a better solution for spectrum and duration conversion in EVC, which will be the focus of this paper. 
%We believe that a unified way of modelling acoustic attributes and speech duration with a seq2seq approach is a better solution for EVC than the frame-based voice conversion frameworks.  
%Recent seq2seq models provides a unified way of modelling acoustic and duration, hence fits better with the EVC task.

%Traditional VC approaches are focused on spectrum mapping with statistical methods, such as Gaussian mixture model (GMM), partial least square regression \cite{helander2010voice} and sparse representation \cite{sisman2019group}. Deep learning approaches, such as deep neural network (DNN) \cite{chen2014voice}, recurrent neural network (RNN) \cite{nakashika2014high} and generative adversarial network (GAN) \cite{sismanstudy} advance the state-of-the-art. These approaches motivate the related studies in EVC.

%Previous EVC studies handle spectrum and prosody together with GMM \cite{tao2006prosody} and sparse representation \cite{aihara2014exemplar}. Recent neural network (NN)-based approaches, such as deep belief network (DBN) \cite{luo2016emotional} and deep bi-directional long-short-term memory network (DBLSTM) \cite{ming2016deep} have also shown their effectiveness. We note that these frameworks require parallel data to train, which is not optimal for real-life applications.

%There have been some NN-based EVC approaches without the need for parallel data, such as cycle-consistent adversarial network (CycleGAN)-based \cite{Zhou2020, shankar2020non} and autoencoder-based frameworks \cite{gao2019nonparallel,zhou2020converting,schnell2021emocat}.
Emotional voice conversion is a special type of voice conversion \cite{sisman2020overview}. Previous studies are focused on frame-based mapping of spectral features of source and target, including using statistical methods~\cite{toda2007voice, sisman2019group} and deep learning methods, such as deep neural network~\cite{chen2014voice}, generative adversarial network (GAN)~\cite{sismanstudy} and CycleGAN \cite{kaneko2017parallel}. Inspired by the success in speaker voice conversion,  these methods are adopted to model both spectral and prosodic parameters for emotional voice conversion. % I HIGHLIGHT THE FRAME-BASED PERSPECTIVE FROM THE FIRST SENTENCE. THIS IS IMPORTANT. 
Successful attempts include GMM~\cite{tao2006prosody}, sparse representation~\cite{aihara2014exemplar}, deep bi-directional long-short-term memory (BLSTM) network \cite{ming2016deep}, GAN-based~\cite{luo2019emotional, Zhou2020,rizos2020stargan,shankar2020non} and auto-encoder-based~\cite{gao2019nonparallel,zhou2020vaw,zhou2020seen,zhou2020converting} methods. These frameworks model the mapping on a frame-by-frame basis. As  emotional prosody is hierarchical in nature~\cite{xu2011speech}, frame-based methods are therefore not the best  in handling prosody conversion~\cite{sisman2020overview}.

Recently, seq2seq models with attention mechanism have attracted much interests in speech synthesis~\cite{kyle2017char2wav,wang2017tacotron} and voice conversion such as SCENT \cite{zhang2019sequence}, AttS2S-VC \cite{tanaka2019atts2s} and ConvS2S-VC \cite{kameoka2020convs2s}. %~\cite{zhang2019sequence, kameoka2020convs2s,zhang2019non} 
%Compared with frame-based methods, seq2seq model allows us to convert the input sequence into an output sequence with different length. 
%\textbf{On the other hand, an unified end-to-end model without vocoder analysis also can lead to better naturalness of synthetic speech~\cite{zhang2019sequence}. [THIS SENTENCE SHOULD NOT BE HERE, IT LOOKS WEIRD AND NOT CONNECTED]} 
Considering that VC and TTS share a similar motivation in a sense that they both aim to generate speech from internal representations~\cite{sisman2020overview}, there are studies to leverage TTS systems to further improve seq2seq VC performance, such as adding text supervision~\cite{zhang2019improving} or leveraging TTS \cite{zhang2019joint,zhang2019non}. Inspired by these studies, seq2seq frameworks have become popular in emotional voice conversion. For example, a seq2seq model is proposed in \cite{robinson2019sequence} to jointly model pitch and duration with parallel data. In~\cite{kim2020emotional}, researchers propose a seq2seq model with multi-task learning for both emotional voice conversion and emotional text-to-speech. We note that these frameworks require tens of hours of emotional speech data to train, which is not practical for real-life scenarios. 

%In this paper, we propose a novel training strategy for seq2seq emotional voice conversion leveraging TTS with limited amount of emotional speech data. 
In this paper, we propose a 2-stage training strategy for seq2seq emotional voice conversion. In stage 1, we perform style initialization, which aims to disentangle speaking style and the linguistic content with a multi-speaker TTS corpus. In stage 2, we perform emotion training, where all components of the network are trained with limited emotional speech data. By doing this, we obtain \textit{emotion encoder} that learns to disentangle emotional style from the speech, and \textit{emotion classifier} that further eliminates the emotion-related information in the linguistic space. The proposed  framework achieves remarkable performance by converting both spectrum and prosody with a limited amount of non-parallel emotional speech data.  %[CAN YOU FINISH THIS SENTENCE? I WANT TO SAY THROUGH THESE WE CAN GENERATE HIGH-QUALITY EMOTIONAL VOICE]} I try to use the style prof. Li talks about in section3. Please read the first paragraph of section 3. Then revise this section

%first pre-train a joint framework of VC and TTS, where \textit{style encoder} learns the speaking style information through from multi-speaker speech data. We then train the framework with the limited amount of emotional speech data, where the style encoder acts as an \textit{emotion encoder} to learn the emotional style information.

The main contributions of this paper include: 1) we propose a seq2seq emotional voice conversion framework leveraging TTS without the need for parallel data, and flexible for many-to-many emotional voice conversion; 2) we propose a novel training strategy that requires a small amount of emotion-labelled data; 3) we significantly improve the performance by modelling the alignment between acoustic and linguistic embedding for emotion styles, which is a departure from frame-based conversion paradigm; 4) 
we propose emotional fine-tuning for WaveRNN vocoder \cite{kalchbrenner2018efficient} training with the limited amount of emotional speech data to further improve the final performance.
%we further show the  improvement on the final performance through WaveRNN vocoder \cite{kalchbrenner2018efficient}.  

This paper is organized as follows: In Section 2, we motivate our study through the comparison with existing seq2seq EVC frameworks. In Section 3, we introduce our proposed framework and the proposed training strategy. In Section 4, we report the experiments. Section 5 concludes the study.

\section{Sequence-to-sequence EVC}

%Sequence-to-sequence model with attention mechanism
%The advent of deep learning has led to a shift from rule-based to data-driven techniques for emotional voice conversion \cite{schuller2020review}.
%Previous EVC studies are focused on modelling the frame-wise mapping of spectral and prosodic speech parameters from emotional speech data \cite{schuller2020review}.
%The seq2seq model with attention mechanism has greatly improved the modelling ability by jointly learning the feature mapping and alignment. It marks the departure from frame-wise modelling to seq2seq modelling \cite{sisman2020overview, schuller2020review}. 

%\textbf{how the 1st sentence is linked to this last sentence? you mention rule based and data-driven, is it related here?} \textbf{Overall, what is the aim of this paragraph? It is not smooth, every sentence is correct but they focus on sth else, not like ONE story. Maybe you can focus on the transition from frame-based to seqtoseq in this paragraph, then it looks better with rest of this section :) }

The seq2seq model, which was first studied in machine translation \cite{bahdanau2015neural}, was found effective in speech synthesis \cite{kyle2017char2wav,wang2017tacotron} and voice conversion \cite{zhang2019sequence, kameoka2020convs2s,zhang2019non,zhang2019joint}. In voice conversion, the seq2seq model with attention mechanism has greatly improved the modelling ability by jointly learning the feature mapping and alignment. The seq2seq model marks a departure from the frame-wise modelling~\cite{sisman2020overview, schuller2020review}.
%We believe that seq2seq modelling would also be a better fit than frame-wise modelling for emotional voice conversion. 
First, the seq2seq model allows for the prediction of the speech duration at the run-time inference which is an essential factor of emotional prosody~\cite{tao2006prosody}. Second, emotion labels are usually annotated at the utterance level in speech corpus~\cite{schuller2020review}, while emotional prosody is supra-segmental and can be associated with only a few words. The attention mechanism makes it possible for the conversion to focus on emotion-relevant regions, which will be our focus. 

There are only a few studies on emotional voice conversion with seq2seq modelling such as jointly modelling pitch and duration with parallel data \cite{robinson2019sequence}, where the output pitch contour is conditioned on the syllable position and source signal; and multi-task learning where a single system is jointly trained for both emotional voice conversion and  text-to-speech~\cite{kim2020emotional}. These frameworks perform well but rely on a large emotional speech corpus. In this paper, we would like to study a limited data solution. To the best of our knowledge, this is the first attempt with the seq2seq model that does not need a large amount of emotional speech training data for EVC. 

\section{Proposed seq2seq EVC model}
%We propose a novel training strategy for seq2seq EVC leveraging TTS that only needs limited amount of emotional speech data. 
%\textcolor{red}{(to Zhou Kun: what is the purpose of defining $\bm{E^t}$, $\bm{E^a}$,  $\bm{E^s}$..., where are the symbols used in this paper? if symbols are not used, we should not have defined them in the first place. It only causes confusions. )}
We propose a seq2seq EVC framework that consists of 5 components, a text encoder, a seq2seq automatic speech recognition (ASR) encoder, a style encoder, a classifier, and a seq2seq decoder. We propose a 2-stage training strategy:  1) Stage I: Style initialization, which disentangles between the speaking style, i.e., speaker style, and the linguistic content with a multi-speaker TTS corpus; 2) Stage II: Emotion training, where all components, initialized by stage I, are further trained with a limited amount of emotional speech data.
%and style encoder acts as emotion encoder to learn the emotion representations with  a limited amount of emotional data. 
Finally, during run-time conversion inference, the framework generates the utterance with reference emotion type by combining the source linguistic representation and the reference emotion representation. While the proposed model is trained to perform both EVC and emotional TTS,  EVC will be the main focus of this paper.
%We first pre-train the proposed framework on a large-scale TTS corpus, where the framework learns to disentangle the speaking style from the linguistic information. We then adapt the framework with limited amount of emotional speech data, where style encoder $E^s$ acts as emotion encoder to learn the emotion representations. At the run-time inference, given a new reference utterance, the framework generates the utterance with reference emotion type by combining the linguistic representation from the source and the emotion representation from the reference. %It is noted that the input of the framework can be either text or audio.
\vspace{-2mm}
\subsection{Training stage I: Style initialization}
%\textbf{please rewrite this section}
At stage I, we adopt a seq2seq VC framework \cite{zhang2019non}, and pre-train it with a publicly available TTS corpus, as shown in Figure \ref{fig:framework}(a). The framework takes the acoustic  features and one-hot phoneme sequences as the inputs. The text encoder and the seq2seq ASR encoder predict the linguistic embeddings from the audio input and the text input respectively. The style encoder embeds the acoustic features into the style embedding. Finally, the seq2seq decoder recovers the acoustic features with the style and linguistic embeddings either from audio or text inputs.

%Given the acoustic features $A$ and one-hot phoneme sequences $T$, the text encoder transforms the text input into linguistic embeddings as $H^t =  E^t(T)$, the seq2seq ASR encoder predicts the linguistic representations of the audio signal as $H^r = E^a(A)$, and the style encoder embeds the acoustic features into a style vector $h^s=E^s(A)$. 
%We then employ an adversarial training to further eliminate the speaker-dependent information in the linguistic representation $H^r$, such as style and speaker information. 
%We use a classifier $C^s$ to predict speaker-dependent information in $H^r$ as $\hat{P}^s = C^s(H^r)$. Given the linguistic and style embeddings, the seq2seq decoder recovers the acoustic features as $\hat{A} = D^a(h^s,H^t)$ or $\hat{A} = D^a(h^s,H^r)$. 
%$M$ and $N$ represent the length of acoustic features and phoneme sequences respectively. 

In stage I, the style encoder learns speaker-dependent information, i.e., speaker style, and excludes linguistic information from the acoustic features. To disentangle from speaker style,  an adversarial training with a classifier
%\textcolor{red}{is it necessary to define $C^s$? is $C^s$ the same as the emotion classifier in Stage 2? or different? the two classifers are very different, one for speaker and another for emotion how can they share the same weights? - see your notes in Figure 2(b) ?} 
is employed to further eliminate speaker information from the linguistic space.  
\begin{figure}[t]
    \centering
    \includegraphics[width=8cm]{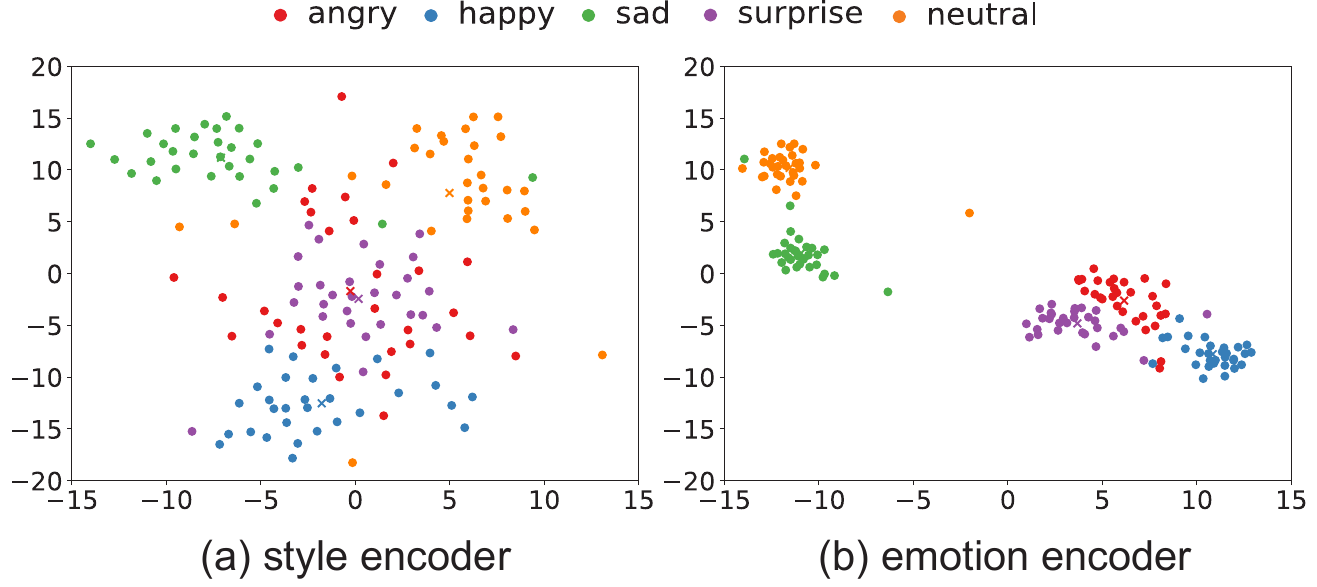}
    \vspace{-7mm}
    \caption{Visualization of emotion embeddings derived from (a) style encoder and (b) emotion encoder. Each point represents the emotion embedding of a reference utterance.} 
    %Each $\times$ symbol represents the mean value of emotion embeddings of all the utterances from the reference set for each emotion.}
    \label{fig:visualization}
    \vspace{-7mm}
\end{figure}
With the text inputs and adversarial training strategy, the framework learns to disentangle the linguistic and style information through a multi-speaker TTS corpus.

However, since the style encoder learns the style information from an emotion-neutral TTS corpus, it does not learn to encode any specific speaking style during stage I, as shown in Figure \ref{fig:visualization}(a). %It may fail
%training samples used in the pre-train are mostly in neutral, we observe that the style encoder fail 
%to provide a clear pattern for different emotion types, as shown in Figure \ref{fig:visualization}(a). 
However, the style encoder has rich knowledge about the style and speaker information, we believe it has the potential to learn the emotional style representation given a small amount of emotional speech data. Therefore, we consider stage I as the style initialization, and propose emotion training in stage II, where the style encoder acts as an emotion encoder to learn the emotional style representations.

%To learn the emotional style representations from the speech, we further adapt the framework with limited amount of emotional speech data, that we will discuss in stage II, 
%where the style encoder acts as the emotion encoder to embed the acoustic features into an emotion vector.
%Therefore, we conduct emotion adaptation in stage II, where the style encoder acts as emotion encoder to learn the emotional style representations.

\begin{figure*}
\vspace{-2mm}
    \centering
    \includegraphics[width=15cm]{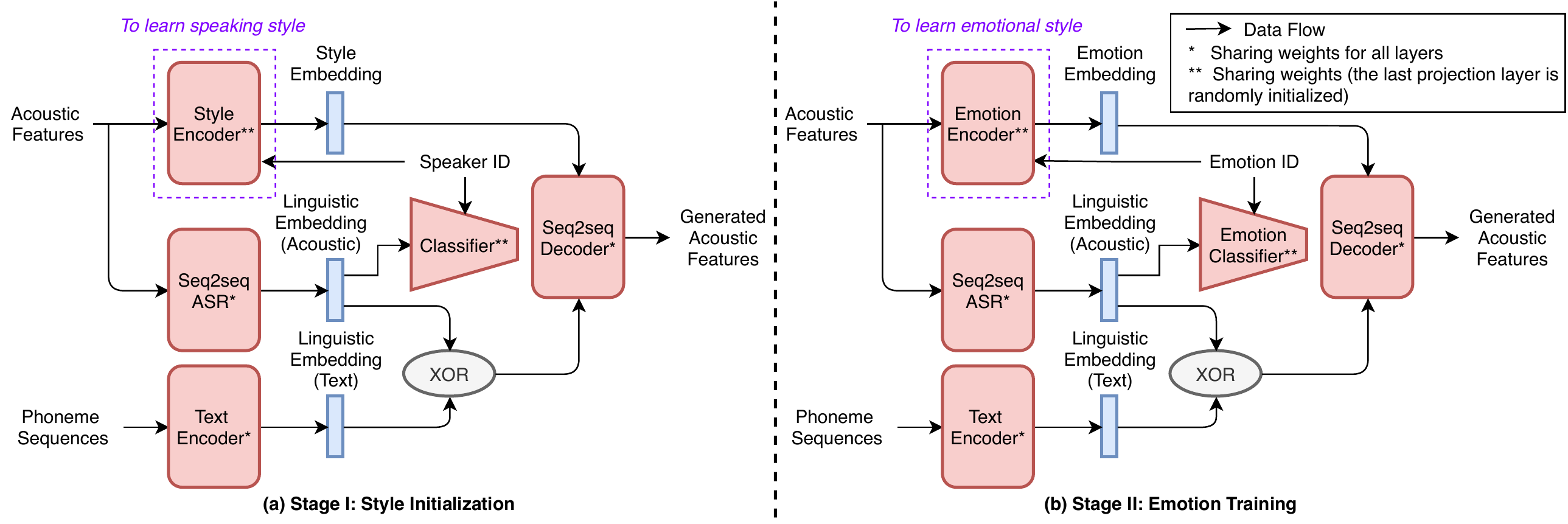}
    \vspace{-3mm}
    \caption{The proposed 2-stage training strategy for seq2seq emotional voice conversion with limited emotional speech data.} %2-stage training for the proposed seq2seq EVC framework leveraging TTS, where the style encoder of stage I acts as emotion encoder in stage II. The pink boxes are involved in the training. XOR denotes exclusive OR operator.}
    \vspace{-6mm}
    \label{fig:framework}
\end{figure*}
\vspace{-2mm}
\subsection{Training stage II: Emotion training}
%\textbf{please rewrite this section, I explained below and in wechat}
We propose to retrain the framework with a limited amount of emotional speech data at stage II, as shown in Figure \ref{fig:framework}(b). We expect that the network has learnt the basic functions of VC and TTS  with a styling mechanism during stage I. The style encoder is then ready to learn the emotional styles from additional emotion-labelled speech data. It acts as an emotion encoder to embed the acoustic features into an emotion vector $h^e$. Furthermore, the classifier acts as an \textit{emotion classifier} to eliminate the emotion information in the linguistic space.   Both the emotion encoder and emotion classifier are trained in a supervised way with a one-hot emotion ID. 
\vspace{-2mm}
\subsubsection{Training with limited emotion data}
The emotion encoder learns the emotional representations through the loss function $L_c$ as below:
\begin{equation}
    L_c = \frac{1}{N}\sum_{n=1}^{N}CE(\phi, \hat{\phi}_n),
\end{equation}
where $CE(\cdot)$ represents the cross entropy loss function, $N$ represents the length of embedding sequence, $\phi$ and $\hat{\phi}_n$ denote the one-hot emotion label and the predicted emotion probability respectively. 
As for the emotion classifier $C^s$, the adversarial loss $L_{adv}$ is modified as follows:
\begin{equation}
    L_{adv} = \frac{1}{N}\sum_{n=1}^{N}\Vert \alpha-\hat{\phi}_n\Vert_2^2,
\end{equation}
where $\alpha = [1/R, ..., 1/R]^T$ is an uniform distribution over the total number of emotion types $R$. And the emotion classification loss $L_{ec}$ is given as:
\begin{equation}
    L_{ec} = CE(\phi, softmax(Vh^e)),
\end{equation}
where $V$ is the weight matrix of the emotion encoder. 

We note that all the components are initialized with the weights learnt at stage I, while the last projection layers of the emotion encoder and the emotion classifier are randomly initialized. %\textcolor{red}{how do we make sure that the huge model is not over-fitted with a small set of data in Stage II if we perform a regular training? usually we fix most parameters and only allow updating of a small set of parameters when we only have a limited amount of data. In this case, we should fix Seq2seq ASR, Text encoder, and Seq2Seq ...}
%By optimizing these three loss functions, the emotion encoder learns the emotional representations from the speech. 
At stage II, we update the entire network during training. The training allows the seq2seq ASR encoder and seq2seq decoder to learn a better alignment between acoustic frames and linguistic embedding sequence that particularly characterizes the emotional style of the utterance. 
%Figure \ref{fig:attention} shows the alignment examples for seq2seq ASR and seq2seq decoder of an utterance. 
Furthermore, we also adapt the speaker style encoder of stage I to an emotion encoder, the speaker classifier of stage I to an emotion classifier.

%\textbf{We note that in the proposed framework XXXXXX mechanism allows us to learn and manipulate duration, which is crucial for emotional voice conversion.}
%\textcolor{red}{The attention mechanism allows us to vary the phonetic duration from source to target during the decoding, which is crucial for EVC. Figure \ref{fig:attention} shows an example of the acoustic-linguistic alignment for (a) converted happy and (b) converted sad utterances. We note that the source utterance is the same for both conversion mappings, while the converted utterances have different duration. These results show that our proposed framework is capable of duration manipulation.}

%both utterances are converted from the same neutral utterance. The acoustic-linguistic alignment and (ii) mel-spectrogram, 
%t shows that our propose framework is capable of converting prosody and duration at the same time.}
%We visualize (i) the alignment example of seq2seq decoder at run-time and (ii) the mel-spectrogram of the converted happy and sad utterances with the same linguistic content.

%\textcolor{red}{(Figure 3 is not very informative, i suggest that we put Figure 3 in the experiment section --- draw 3 plots, one from source emotion ( happy for example), second from target reference emotion (angry for example), and right for generated target (angry for example) , in this way, we show that we can convert happy prosody to angry sucessfully. }% As we mainly focus on EVC here, we will study the text encoder further as future work.

Overall, the framework leverages the knowledge of disentanglement between linguistic and style information learnt at stage I, and effectively learns the emotional style disentanglement only with a limited amount of emotional speech data at stage II. The proposed 2-stage training further helps with obtaining better disentangled emotional representation without the support of large emotional speech data.

%\textbf{By optimizing these loss functions [what do you mean these loss functions? only 3-4 loss functions??]}, 

%The emotion encoder $E^s$ learns to disentangle the emotional style representations from the speech. And the emotion classifier can further eliminate the emotion-related information in the linguistic space to achieve a better disentanglement between linguistic and emotional information.

%\textbf{Can you mention in 3-4 lines here about what the proposed idea do, and how it works? I mean the WHOLE framework. Not just emotion encoder. Because this stage aims to generate emotional speech, right (both VC and TTS)?. So you should mention this as an overall summary here. You should not end this part with emotion classifier. You should end it with the whole "emotional VC" perspective and how your idea allows us to get emotional speech. At the end you propose Emotional VC that uses the emotion encoder etc. I think these 2 sections are not well-written (English is OK, but the story is missing; the readers cannot understand the proposed idea or framework), you need to go through everything till here to make story clear.  The readers should understand how your training allows us to get seqtoseq emotional VC with limited data. Limited data MUST be mentioned here too. This part should be a nice ending about your framework. You may start with "Overall, "} 
\vspace{-3mm}
\subsubsection{Style encoder vs. emotion encoder}

During the emotion training, the style encoder acts as an emotion encoder and takes emotion ID as the input to effectively learn the emotion representation in the speech. To validate our idea, we use t-SNE \cite{maaten2008visualizing} to visualize the emotion embedding of the reference utterances, which are derived by the style encoder from stage I and the emotion encoder from stage II respectively.
To our delight, as shown in Figure \ref{fig:visualization}, the emotion embeddings derived by the emotion encoder form separate groups for each emotion type, while those from the style encoder fail to provide a clear pattern.
From  Figure \ref{fig:visualization}, we also observe a significant separation between the emotions with lower values of arousal and valence such as neutral and sad, and those with higher values such as angry, happy and surprise. These observations further validate our 2-stage training for EVC.

\vspace{-3mm}
\subsection{Run-time inference}
%\textbf{some sentences need to change place, please check below:}
At run-time, we use the emotion encoder to generate the emotion embeddings from a set of reference utterances belonging to the same emotion category. We use the average emotion embedding to represent the emotion style.  %It is noted that the reference set has no overlap with the evaluation utterances.
Given a source utterance and the intended emotion category, we use a seq2seq ASR encoder to derive the linguistic embedding of the source utterance, and apply the respective emotion embedding to the decoder. The converted acoustic features can be reconstructed by the seq2seq decoder. 

\vspace{-3mm}
\subsection{Comparison with related work}
The proposed seq2seq EVC framework shares a similar motivation with \cite{kim2020emotional,zhang2019non} in terms of leveraging TTS but differs in many aspects. 
%To learn a better emotion disentanglement, we employ adversarial training with an emotion classifier to further eliminate the emotion information in the linguistic embedding; and 3) in order to improve the intelligibility of the converted speech, we use a seq2seq recognition encoder, and add an explicit loss between the embedding of VC and TTS in the linguistic space to get a better alignment. 
To start with, \cite{zhang2019non} only focuses on speaker disentanglement, and emotion has not been considered. The proposed 2-stage training strategy allows the network to learn emotion style in training stage II, thus, requires a smaller amount of emotional speech data. %Furthermore, we propose to learn the prosody disentanglement as style initialization, and emotion disentanglement during the emotion training. 
Compared with~\cite{kim2020emotional} which needs more than \textit{30 hours} of emotional speech data for training, our proposed framework only uses \textit{less than 50 minutes} of emotional speech data. Besides, we further employ adversarial training with an emotion classifier to learn a better emotion disentanglement and use a seq2seq ASR with an explicit loss between the linguistic embedding of EVC and TTS to get a better alignment.
\vspace{-3mm}
\section{Experiments}
We conduct emotion conversion from neutral to angry, sad, happy and surprise, denoted as \textit{Neu-Ang}, \textit{Neu-Sad}, \textit{Neu-Hap}, and \textit{Neu-Sur} respectively. We first use VCTK corpus \cite{veaux2016vctk} for stage I as shown in Figure 2(a), and then use the ESD database~\cite{zhou2021emotional} for stage II as shown in Figure 2(b). For each emotion pair, we use 300 utterances for training, 30 utterances for reference, and 20 utterances for evaluation. The total duration of emotional speech data used in the stage II is around \textit{50 minutes}, which is small in the context of seq2seq training. 
%Our proposed framework shares a similar motivation with the one in \cite{zhang2019non} but differs in many ways. In \cite{zhang2019non}, researchers propose to learn the speaker disentanglement, but ours propose to learn the prosody disentanglement during the pre-train, and the emotion disentanglement during the adaptation. In \cite{zhang2019non}, the speaker encoder is frozen during the fine-tune. In our method, we propose to adapt the style encoder as the emotion encoder to learn the emotion representation during the adaptation. Our proposed training strategy also provides a solution for emotional voice conversion studies when we do not have the sufficient emotional speech data.

\begin{table}[t]
\centering
\caption{A comparison of MCD {[}dB{]} values.}
\vspace{-3mm}
\scalebox{0.65}{
\begin{tabular}{c|cccc}
\hline
\multirow{2}{*}{Framework} & \multicolumn{4}{c}{MCD {[}dB{]}}      \\ \cline{2-5} 
                           & Neu-Ang & Neu-Sad & Neu-Hap & Neu-Sur \\ \hline
CycleGAN-EVC \cite{Zhou2020}               & 4.57    & 4.32    & 4.46    & 4.68    \\ 
StarGAN-EVC \cite{rizos2020stargan}                & 4.51    & 4.31    & 4.24    & 4.39    \\
Baseline Seq2seq-EVC & 5.14    & 5.27   &  5.04  & 5.40   \\ 
Seq2seq-EVC-GL           & 3.98    & 3.83    & 3.92    & 3.94    \\ 
Seq2seq-EVC-WA1     & 3.72    & 3.73    & 3.71    & 3.83    \\ 
Seq2seq-EVC-WA2      & 3.73    & 3.73    & 3.70    & 3.80    \\ \hline
\end{tabular}}
\label{tab:mcd}
\vspace{-5mm}
\end{table}

%All the speech data is sampled at 16 kHz. We use 80-dimensional log Mel-spectrograms extracted every 12.5 ms with the STFT of 50 ms as the acoustic features. We use the phoneme transcription as the input to the text encoder. 

The codes and implementation details of this work are publicly available at: \url{https://github.com/KunZhou9646/seq2seq-EVC}. We implement two state-of-the-art methods, together with 4 seq2seq EVC systems:
\begin{itemize}
    \item CycleGAN-EVC \cite{Zhou2020} (\textit{baseline}): CycleGAN-based emotional voice conversion with WORLD vocoder. 
    \item StarGAN-EVC \cite{rizos2020stargan} (\textit{baseline}): StarGAN-based emotional voice conversion with WORLD vocoder. 
    \item Baseline Seq2seq-EVC (\textit{baseline}):  Seq2seq-EVC trained directly with limited ESD data without any pre-training, and followed by a  Griffin-Lim vocoder \cite{griffin1984signal};
    \item Seq2seq-EVC-GL (\textit{proposed}):  Seq2seq-EVC followed by a  Griffin-Lim vocoder;
    \item  Seq2seq-EVC-WA1 (\textit{proposed}):  Seq2seq-EVC followed by a  WaveRNN vocoder \cite{kalchbrenner2018efficient} that is pre-trained on VCTK corpus;
    \item  Seq2seq-EVC-WA2 (\textit{proposed}): Seq2seq-EVC followed by a  WaveRNN vocoder that is pre-trained on VCTK corpus, and fine-tuned with limited ESD data.
\end{itemize}
We note that CycleGAN-EVC only can perform the one-to-one conversion, thus we train one CycleGAN-EVC for each emotion pair separately. Both StarGAN-EVC and our proposed Seq2seq-EVC use a unified model for all the emotion pairs.
%The codes and speech samples are publicly available\footnote{\textbf{Codes \& Speech Samples}: \url{https://xxx}}.

%\begin{tabular}[c]{@{}c@{}} Baseline Seq2seq-EVC \\ (randomly initialized)\end{tabular}
\begin{table}[t]
\centering
\caption{A comparison of DDUR {[}s{]} values for the voiced parts.}
\vspace{-3mm}
\scalebox{0.65}{
\begin{tabular}{c|cccc}
\hline
\multirow{2}{*}{Framework} & \multicolumn{4}{c}{DDUR {[}s{]}}     \\ \cline{2-5} 
                           & Neu-Ang & Neu-Sad & Neu-Hap & Neu-Sur \\ \hline
Source-Target              & 0.36    & 0.46    & 0.26    & 0.44    \\ 
 Baseline Seq2seq-EVC  & 0.65   & 0.91  &  0.69  &  0.54  \\ 
Seq2seq-EVC-GL                & 0.38    & 0.41    & 0.26    & 0.33    \\
Seq2seq-EVC-WA1                & 0.39    & 0.39    & 0.27    & 0.33    \\
Seq2seq-EVC-WA2                & 0.34    & 0.40    & 0.24    & 0.32    \\\hline
\end{tabular}}
\label{tab:ddur}
\vspace{-3mm}
\end{table}

\begin{figure}[t]
\centering
\subfigure[Converted Happy]{
\begin{minipage}[c]{0.5\linewidth}
\centering
\includegraphics[width=3cm]{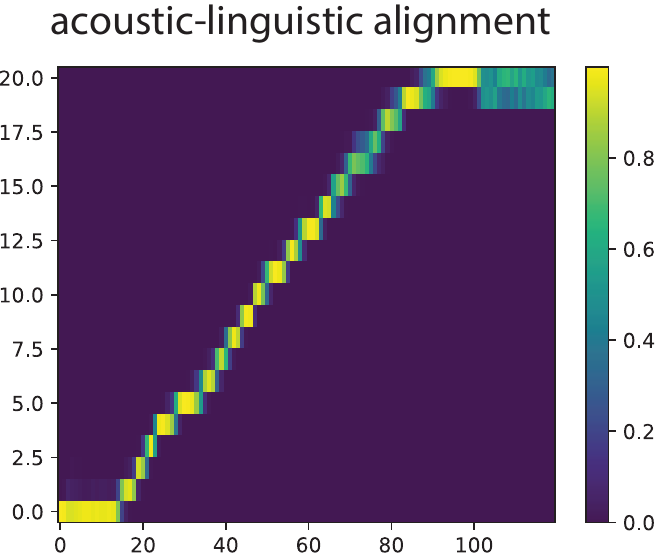}
\end{minipage}%
}%
\centering
\subfigure[Converted Sad]{
\begin{minipage}[c]{0.5\linewidth}
\centering
\includegraphics[width=3cm]{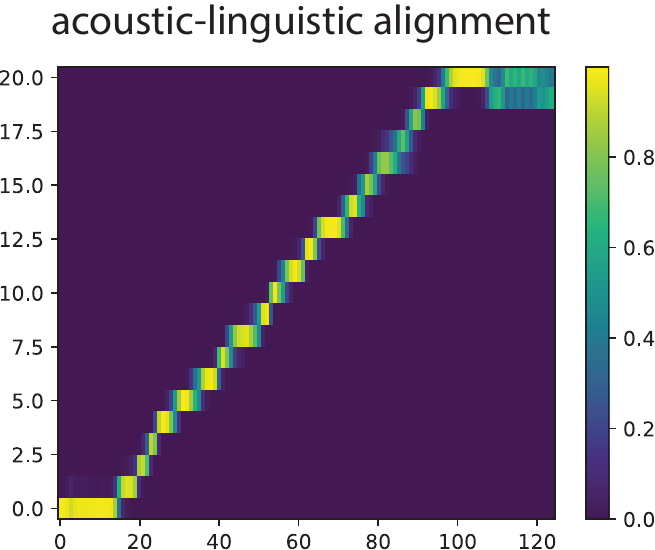}
\end{minipage}%
}%
\centering
\vspace{-5mm}
\caption{Acoustic-linguistic alignment visualization of utterances that are converted from neutral to (a) happy and (b) sad. } 
\vspace{-7mm}
\label{fig:attention}
\end{figure}
%\subsection{Experimental Setup}

%The text encoder is 3-layer of 1D CNN with a kernel size of 5 and the channel of 512, followed by 1-layer of  256-cell BLSTM and a fully connected (FC) layer with the output channel of 512. The seq2seq ASR encoder consists of an encoder which is a  2-layer 256-cell BLSTM, and a decoder which is a  1-layer 512-cell LSTM with an attention layer and followed by a FC layer with the output channel of 512. The style encoder is 2-layer of 128-cell BLSTM followed by a FC layer with the output channel of 128. The classifier is 4-layer of FC with the channel of \{512, 512, 512, 99\}. The seq2seq decoder follows the same model architecture of the one used in Tacotron \cite{wang2017tacotron}. During pre-training, we set the learning rate as 0.0001 for 200 epochs. For the emotion adaptation, we set the learning rate as 0.0001 and half it every 7 epochs. We set the batch size as 64 and 32 for pre-training and adaptation respectively. The WaveRNN vocoder predicts 9-bits waveform with $\mu$-law companding. Its implementation follows a publicly available version\footnote{\url{https://github.com/fatchord/WaveRNN}}.  
%\vspace{-2mm}
\subsection{Objective Evaluation}
\vspace{-1mm}
We calculate Mel-cepstral distortion (MCD) \cite{sisman2020overview} and the average absolute differences of the utterance duration (DDUR) \cite{zhang2019non} for the voiced parts to measure the spectral distortion and duration difference respectively. 
In Seq2seq-EVC models, Mel-spectrograms are adopted as acoustic features, and the Mel-cepstral coefficients (MCEPs) are extracted directly from the waveform to calculate MCD values. 

%\textbf{For Seq2seq-EVC models, we calculate the MCD values on the MCEPs that are calculated from the waveform directly.}
To motivate our proposed 2-stage training, we first conduct experiments with the Baseline Seq2seq-EVC model that is trained directly with limited ESD data without any pre-training procedures. We note that in all experiments, it consistently achieves the worst results in terms of MCD and DDUR values. This observation shows that seq2seq-EVC does not work well  with limited training data, which further shows the necessity of our proposed 2-stage training strategy.
%for EVC under limited emotional speech data.

As shown in Table \ref{tab:mcd}, our proposed Seq2seq-EVC models with Griffin-Lim, WaveRNN and fine-tuned WaveRNN always outperform baseline CycleGAN-EVC and StarGAN-EVC. We also note that the proposed Seq2seq-EVC-WA1 and Seq2seq-EVC-WA2 consistently achieve the best results of MCD values for all the emotion pairs, which shows the effectiveness of our proposed 2-stage training strategy for limited data EVC.

%We then calculate the average absolute differences of the utterance duration (DDUR) for the voiced parts between the converted and the target utterances. 
We note that the attention mechanism allows us to vary the phonetic duration from source to target during the decoding, which is crucial for EVC. Figure \ref{fig:attention} shows an example of the acoustic-linguistic alignment for (a) converted happy and (b) converted sad utterances. We note that the source utterance is the same for both conversion mappings, while the converted utterances have different duration. These results further show that our proposed framework is capable of duration manipulation.

We further report DDUR results to evaluate duration conversion performance in Table \ref{tab:ddur}. We observe that the proposed Seq2seq-EVC-WA1 (with WaveRNN) and Seq2seq-EVC-WA2 (with fine-tuned WaveRNN) consistently achieves the best DDUR results for all emotion pairs. We noted that baseline frameworks CycleGAN-EVC and StarGAN-EVC cannot modify speech duration, hence they are not reported in the table. These results show the effectiveness of our proposed Seq2seq-EVC framework in terms of duration conversion.

\begin{figure}[t]
    \centering
    \includegraphics[width=7cm]{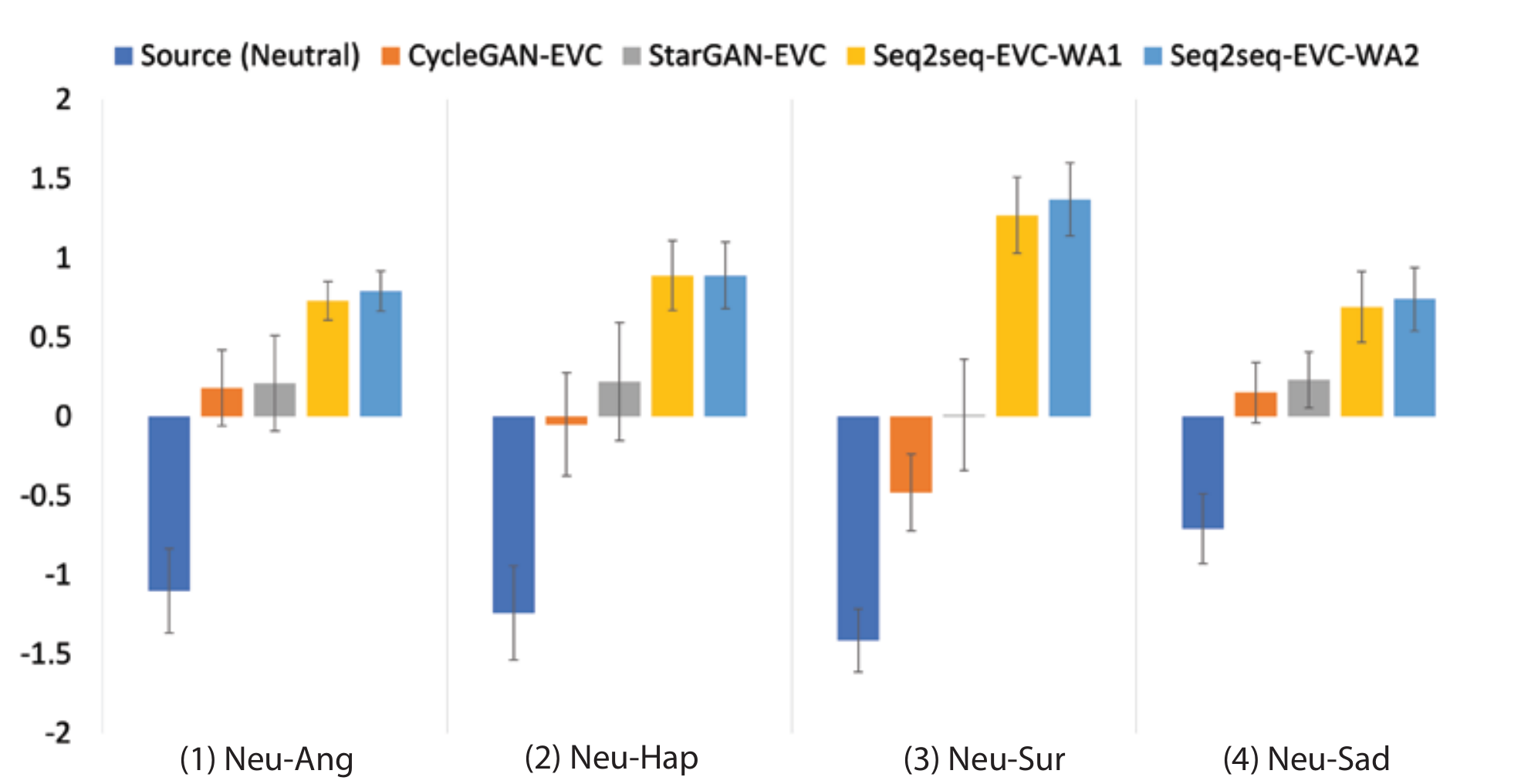}
    \vspace{-3mm}
    \caption{Emotional similarity results with 95\% confidence interval to evaluate emotion similarity with target speech in a scale of -2 to 2 (-2: absolutely different; -1: different; 0: cannot tell; +1: similar; +2: absolutely similar).}
    \label{fig:mos}
    \vspace{-3mm}
\end{figure}

\begin{table}[t]
\caption{Best Worst Scaling (BWS) listening experiments to evaluate the overall speech quality.}
\vspace{-3mm}
\centering
\scalebox{0.6}{
\begin{tabular}{c|c|ccc}
\hline
\multicolumn{2}{c|}{Systems}  & Seq2seq-EVC-GL & Seq2seq-EVC-WA1 & Seq2seq-EVC-WA2 \\ \hline
\multirow{2}{*}{Neu-Ang} & Best  & 0\%            & 19\%            & \textbf{81\%}            \\ 
                         & Worst & 94\%           & 6\%             & \textbf{0\%}             \\ \hline
\multirow{2}{*}{Neu-Hap} & Best  & 0\%            & 32\%            & \textbf{68\%}            \\ 
                         & Worst & 97\%           & 3\%             & \textbf{0\%}             \\ \hline
\multirow{2}{*}{Neu-Sur} & Best  & 6\%            & 25\%            & \textbf{69\%}            \\ 
                         & Worst & 94\%           & \textbf{3\%}            & \textbf{3\%}             \\ \hline
\multirow{2}{*}{Neu-Sad} & Best  & 0\%            & 10\%            & \textbf{90\%}            \\ 
                         & Worst & 94\%           & 6\%             & \textbf{0\%}             \\ \hline
\end{tabular}}
\label{tab:bws}
\vspace{-6mm}
\end{table}
%\begin{table}[t]
%\caption{Best Worst Scaling (BWS) listening experiments to evaluate the overall speech quality.}
%\scalebox{0.7}{
%\begin{tabular}{|c|c|c|c|c|c|c|c|c|}
%\hline
%\multirow{2}{*}{Systems} & \multicolumn{2}{c|}{Neu-Ang} & \multicolumn{2}{c|}{Neu-Hap} & \multicolumn{2}{c|}{Neu-Sur} & \multicolumn{2}{c|}{Neu-Sad} \\ \cline{2-9} 
%                         & Best         & Worst         & Best         & Worst %        & Best         & Worst         & Best         & Worst         \\ \hline
%Seq2seq-EVC-GL           & 0\%          & 94\%          & 0\%          & 97\%          & 6\%          & 94\%          & 0\%          & 94\%          \\ \hline
%Seq2seq-EVC-WA1          & 19\%         & 6\%           & 32\%         & 3\%           & 25\%         & 3\%           & 10\%         & 6\%           \\ \hline
%Seq2seq-EVC-WA2          & 81\%         & 0.0\%         & 68\%         & 0\%           & 69\%         & 3\%           & 90\%         & 0\%           \\ \hline
%\end{tabular}}
%\end{table}

\vspace{-2mm}
\subsection{Subjective Evaluation}

We conduct listening tests to assess the emotion similarity and speech quality. 15 subjects participated in all the experiments and each listened to 128 converted utterances in total.

We first report the emotion similarity results as shown in Figure \ref{fig:mos}. We use the baseline frameworks CycleGAN-EVC and StarGAN-EVC; and the proposed framework Seq2seq-EVC with Griffin-Lim (Seq2seq-EVC-GL), WaveRNN (Seq2seq-EVC-WA1), and fine-tuned WaveRNN (Seq2seq-EVC-WA2). All participants are asked to listen to the reference target speech first, and then score the speech samples in terms of the emotion similarity to the reference target speech.
%in a scale of -2 to +2 (-2: absolutely different; -1: different; 0: cannot tell; +1: similar; +2: absolutely similar). 
It is encouraging to see that the proposed Seq2seq-EVC framework with WaveRNN (Seq2seq-EVC-WA1) and fine-tuned WaveRNN (Seq2seq-EVC-WA2) significantly outperform the baselines for all the emotion pairs, especially for Neu-Sur. % These promising results show the effectiveness of our proposed Seq2seq-EVC framework in terms of emotion conversion.

We further conduct the best-worst scaling (BWS) \cite{ccicsman2017sparse} test in terms of speech quality of our proposed Seq2seq-EVC framework with 1) Griffin-Lim (Seq2seq-EVC-GL), 2) WaveRNN (Seq2seq-EVC-WA1), and 3) fine-tuned WaveRNN (Seq2seq-EVC-WA2). All participants are asked to choose the best one and the worst one in terms of the overall quality. From Table \ref{tab:bws}, Seq2seq-EVC-WA2 outperforms
the baseline consistently, which proves the effectiveness of our emotional fine-tuning strategy on WaveRNN vocoder. 
\vspace{-3mm}
\section{Conclusion}
In this paper, we propose a novel training strategy for seq2seq emotional voice conversion leveraging text-to-speech without the need for parallel data. 
To our best knowledge, this is the first work of seq2seq emotional voice conversion that only needs a limited amount of emotional speech data to train. Moreover, the proposed framework can do many-to-many emotional voice conversion, and conduct spectral and duration mapping at the same time. We also investigate the training strategy of emotion fine-tuning for WaveRNN vocoder training.
%Moreover, the proposed framework does not need parallel data and can do many-to-many emotional voice conversion. 
Experimental results show a significant improvement of the conversion performance over the baselines. 
\vspace{-3mm}
\section{Acknowledgment}
The research is funded by SUTD Start-up Grant Artificial Intelligence for Human Voice Conversion (SRG ISTD 2020 158) and SUTD AI Grant - Thrust 2 Discovery by AI (SGPAIRS1821), the National Research Foundation, Singapore under its AI Singapore Programme (Award No: AISG-GC-2019-002) and (Award No: AISG-100E-2018-006), and its National Robotics Programme (Grant No. 192 25 00054), and by RIE2020 Advanced Manufacturing and Engineering Programmatic Grants A1687b0033, and A18A2b0046.
%\bibliographystyle{IEEEtran}

%\bibliography{mybib}
\bibliographystyle{IEEEtran}
{\footnotesize
\bibliography{mybib}
}

\end{document}